\documentclass[conference]{IEEEtran}
\usepackage{cite}
\usepackage{amsmath,amssymb,amsfonts}
\usepackage{algorithmic}
\usepackage{graphicx}
\usepackage{textcomp}
\usepackage{xcolor}
\usepackage{booktabs}
\usepackage{hyperref}
\usepackage{url}

\newif\iffinal
\finaltrue

\begin{document}

\title{Classification of Inkjet Printers based on Droplet Statistics
}

\iffinal
    \author{\IEEEauthorblockN{1\textsuperscript{st} Patrick Takenaka}
    \IEEEauthorblockA{
            \textit{Institute for Applied Artificial Intelligence}\\
            \textit{Stuttgart Media University}\\
            Stuttgart, Germany \\
            takenaka@hdm-stuttgart.de
            }
\and
\IEEEauthorblockN{2\textsuperscript{nd} Manuel Eberhardinger}
\IEEEauthorblockA{
            \textit{Institute for Applied Artificial Intelligence}\\
            \textit{Stuttgart Media University}\\
            Stuttgart, Germany \\
            eberhardinger@hdm-stuttgart.de
            }
\and
\IEEEauthorblockN{2\textsuperscript{nd} Daniel Grie\ss{}haber}
\IEEEauthorblockA{
            \textit{Institute for Applied Artificial Intelligence}\\
            \textit{Stuttgart Media University}\\
            Stuttgart, Germany \\
            griesshaber@hdm-stuttgart.de
            }
\and
\IEEEauthorblockN{3\textsuperscript{rd} Johannes Maucher}
\IEEEauthorblockA{
            \textit{Institute for Applied Artificial Intelligence}\\
            \textit{Stuttgart Media University}\\
            Stuttgart, Germany \\
            maucher@hdm-stuttgart.de
            }
}   
\fi

\maketitle

\begin{abstract}
Knowing the printer model used to print a given document may provide a crucial lead towards identifying counterfeits or conversely verifying the validity of a real document. Inkjet printers produce probabilistic droplet patterns that appear to be distinct for each printer model and as such we investigate the utilization of droplet characteristics including frequency domain features extracted from printed document scans for the classification of the underlying printer model. We collect and publish a dataset of high resolution document scans and show that our extracted features are informative enough to enable a neural network to distinguish not only the printer manufacturer, but also individual printer models.
\end{abstract}


\section{Introduction}
Printing text on paper has been the basis for communication and the dissemination of information for a long time. While nowadays many paper-driven processes are gradually digitized, they still play a crucial role in daily life. At the same time, new technologies make it easy for non-experts to use high-quality printing methods to forge or alter documents. These technologies include inexpensive printers and scanners, various image processing software and also tools supported by artificial intelligence \cite{kelly_forensic_2020}. The incentive for criminal activity is high, as the entry barrier for interested individuals is low and there are many different types of possible forgeries, such as concert tickets, travel tickets, banknotes, fake IDs or other kinds of personal documents. Additionally, it is hard for non-experts to decide if the presented documents are genuine or forged. These points make it difficult for law enforcement authorities and also for counterfeit detection in banks to keep up with the latest counterfeits \cite{auberson_development_2016}. Identifying the printer model that was used to produce a particular document is an important aspect for determining whether it is genuine or fake, as in many circumstances the printer model for genuine documents is known and thus can be compared with.

For laser printers, there is already a solution for recognizing which printer has created a document by decoding the machine identification code, which is displayed as yellow dots on the document and is not visible to the naked eye. For inkjet printers, however, there is still no general method applicable without the help of spectroscopy \cite{kumar_spectroscopic_2020, al-ameri_spectral_2022, asri_discrimination_2021, gal_principal_2015}. In this work, we present a method that works only with digital image data from high-quality scans, without the need for the original document or additional specialized hardware, such as Raman Spectrometers. Machine learning, and especially deep learning, often requires large amounts of labeled data, which in the case of printer identification is challenging due to the logistical challenges of gathering many high-resolution scans from various printers. To mitigate this problem, we introduce a carefully designed feature extraction from image data and show that this leads to significantly better classification performance than relying on image features alone. Our core assumption is that each printer model exhibits distinct droplet patterns, and by extracting features based on the frequencies in those patterns we can capture their identity. Based on insights from domain experts we continue by also adding features related to droplet shapes specifically, such as their areas and boundaries, and thus arrive at a feature vector that both captures the patterns and the visual aspects as well. 

\begin{figure}[t]
    \centering
    \includegraphics[width=.5\textwidth]{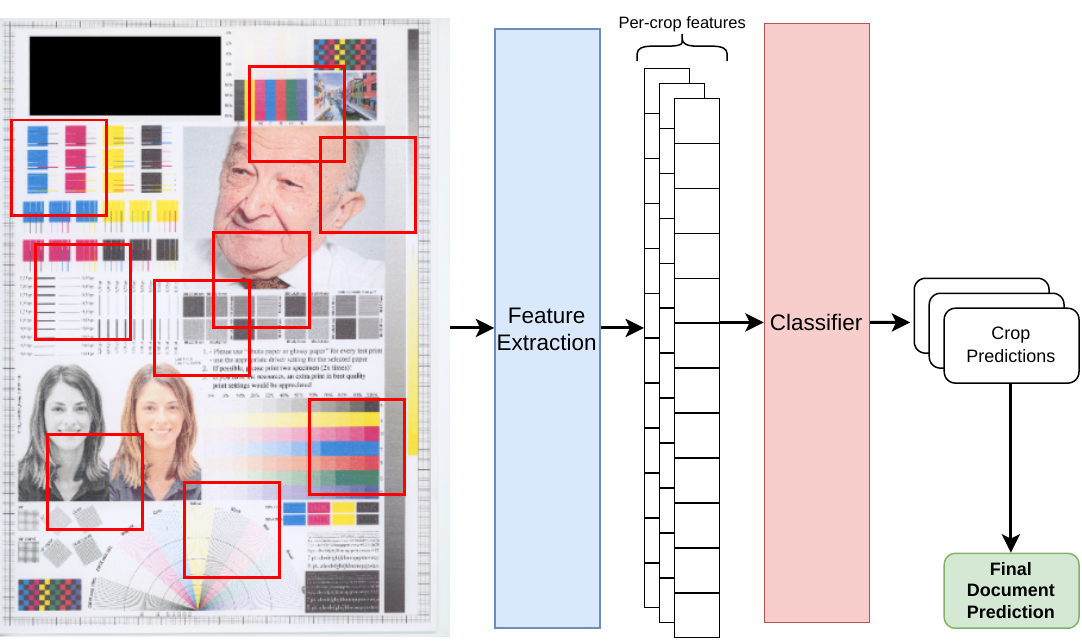}
    \caption{Overview of our printer identification pipeline. Features are extracted in parallel from random crops of the given document, after which they are fed to a classifier model. Each crop is processed individually and their corresponding predictions aggregated to produce a final class prediction for the whole document.}
    \label{fig:overview}
\end{figure}

We further introduce a training and evaluation scheme that can handle the large amounts of data present in a high-resolution scan in an efficient manner and verify it empirically. An overview of the overall classification process is shown in Fig.~\ref{fig:overview}.

We further collect and introduce a dataset for printer identification, which consists of 50 high-resolution scans of documents printed from 25 different printer models across multiple manufacturers. To the best of our knowledge there is no other dataset in this scale publicly accessible for printer identification. Our dataset and code is available at \url{https://github.com/P-Takenaka/ijcnn2024-inkjet-classification}.

Our objective is to provide the foundation for a printer identification tool with minimal hardware and expert knowledge requirements. As such, our contributions in this work are as follows:

\begin{itemize}
    \item Introduction of a new dataset for printer identification
    \item A carefully designed feature set for printer identification extracted from image data
    \item A training and evaluation scheme in order to handle the high-resolution image data
\end{itemize}

The rest of this paper is structured as follows: We first give a brief overview of related work on printer identification in Sec.~\ref{section:related-work}. We continue by providing the necessary background knowledge w.r.t. the frequency-based features and general ink-jet printing technology in Sec.~\ref{section:background}, followed by the presentation of the dataset and the feature extraction method in Sec.~\ref{section:data}. The design of the experiments and the evaluation of the printer classification method is described in Sec.~\ref{section:experiments}. We finally discuss the limitations and future work in Sec.~\ref{section:limitations}.

\section{Related Work}
\label{section:related-work}
The task of identifying or distinguishing printers has long been of great interest \cite{zimmerman_preliminary_1986, cantu_analytical_1991, totty_examination_1990, gilmour_comparison_1994, merrill_forensic_2003} and is usually done with spectroscopy in combination with machine learning approaches \cite{kumar_spectroscopic_2020, al-ameri_spectral_2022, asri_discrimination_2021, gal_principal_2015}. Another approach is using textual features based on the differences in the geometric distortion of printed fonts \cite{bulan_geometric_2009, joshi_source_2020, joshi_source_2022, escher_robustness_2017}. 
Other work focuses on detecting if barcodes were printed from the same source \cite{guo_printer_2024}. In our work, however, we do not constrain the image regions---to for instance barcodes---used for classification in order to allow better generalization across a wide variety of documents.
A similar work based on feature engineering has been studied in \cite{gaubatz_printer-scanner_2009}, where a feature vector is computed over different quality assurance metrics such as color variations, but the study is limited to only four printers and four scanners. In \cite{tsai_digital_2013}, the discrete wavelet transform was used to extract features from printed documents to classify the source printer for Chinese characters. In \cite{tsai_digital_2018, tsai_deep_2019}, several different feature extraction methods were studied and compared, focusing only on laser printers. Our work, in contrast, focuses on inkjet-printers.

In contrast to printer identification, there is also research aimed at anonymizing documents \cite{richter_forensic_2018} or using Siamese neural networks to predict whether two documents were printed by the same printer \cite{ferreira_ensembling_2021}.

Frequency-based features are often used to reduce the difficulty of the problem for machine learning algorithms. The Fourier transform is used in the field of medical imaging to improve the performance of deep learning models \cite{qi_accurate_2021, juntunen_deep-learning-assisted_2022, yue_fourier_2020} or for fraud detection in manufacturing plants \cite{mohammad-alikhani_fault_2023, benkedjouh_deep_2018}. Wavelet features are widely used across all domains, e.g. for medical purposes \cite{akut_wavelet_2019, sarhan_brain_2020, savareh_wavelet-enhanced_2019, serte_wavelet-based_2020}, for high-resolution image reconstruction \cite{chen_wavelet_2020}, for seismic data reconstruction \cite{liu_seismic_2022} or the prediction of emissions into the atmosphere \cite{qiao_forecasting_2019}.

\section{Background}
\label{section:background}
Here we first describe the necessary background for frequency domain transformation, and---in order to give a better picture of the intuition behind our feature extraction---continue by illustrating the general printing process prevalent in inkjet-printing.

\subsection{Frequency Domain Features}

In the following we describe the theory of different established methods for transforming an---possibly multi-dimensional---input signal into the frequency domain.

\textbf{Fourier Analysis:}
The Fourier transform converts an input signal from the temporal- or, in our case spatial domain into frequency domain by deconstructing the signal into a series of (possibly infinite) sine and cosine waves. For discrete, equally spaced signals the discrete Fourier transform (DFT) therefore is a Z-transform that converts the signal $x(t)$ of length $T$ into $T$ complex coefficients so that 
\begin{equation}
    X(k) = \sum_{t=0}^{T-1}{x(t)\cdot e^{-i2\pi \frac{k}{T}t}}
\end{equation}
where $X(k)$ are polar coordinates defining the phase and amplitude of the sinusoidal component of the signal $x(t)$ with frequency $\frac{k}{N}$.
The resulting coefficients of the Fourier transform therefore describe global frequency information of the input signal.

\textbf{Wavelet Transform:}
While the DFT, and Fourier analysis in general, have the limitation to only capture global frequency information, wavelet transformations are able to preserve the temporal or spatial locality information while also capturing the spectral transformation of the signal. The discrete wavelet transformation (DWT) of a one-dimensional signal $x(t)$ can be described as
\begin{equation}
    W_\psi(m, n) = \sum_{t=0}^{T-1}{\psi_{m, n}(t)\cdot x(t)}
\end{equation}
where $\psi_{m, n}(t)$ is a \textit{mother-wavelet} function that is scaled by $m$ and translated by $n$, so that $W_\psi(m, n)$ describes the correlation of the signal $x(t)$ with this transformed wavelet function. Since the wavelet function is limited in duration, the shifting with $n$ causes the calculated correlation to describe the response in a narrow window of the signal, while the scaling using $m$ causes a frequency dependent response.

\textbf{Short-time Fourier Transform}
The short-time Fourier transform (STFT) is an alternative approach for extracting spectral information from a signal while maintaining temporal locality. This is achieved by a sliding windowing function that effectively masks out chunks of the signal. Each chunk is then transformed into frequency domain independently:
\begin{equation}
    X_w(k, n) = \sum_{t=0}^{T-1}{w_n(t)\cdot x(t)\cdot e^{-i2\pi \frac{k}{T}t}}
\end{equation}
where $w_n(t)$ is the windowing function shifted by $n$.

While the STFT contains the real frequency information of the signal (compared to the correlation with the wavelet function in the DFT), the width of the windowing function trades of the temporal and frequency resolution, while the DFT does not need such a trade-off.

\subsection{Inkjet Printing} Inkjet printers construct an image by placing droplets of colored ink onto a medium. The droplets can be produced by different methods such as thermal expansion or mechanical stimulation using piezo electric devices \cite[p. 58 ff.]{hoath_fundamentals_2016}. In the most common setting that is of interest here, the droplets are ejected onto a white paper medium, so that the density of droplets can be used to perform subtractive color mixing to create arbitrary colors by only using a few primary colors, most commonly magenta, cyan and yellow. For text and other black and white printouts, often there is an additional black ink to avoid the need to mix this common color. Due to the limited resolution of the human eye, areas of small dots, even if non-overlapping, will be perceived as a single color depending on the density of dots of each primary color. 

The size of dots produced by inkjet printers typically varies between $10-100\mu{}m$\cite[p. 10]{hoath_fundamentals_2016}. To produce drops of this size with the required accuracy and frequency, basic physical properties of the used ink need to be considered such as surface tension and viscosity. Since typically printing is done in multiple lines, the ink nozzles and medium move relative to another, requiring planning of acceleration and timing of drop ejection.
Due to this complexity, the position and size of the individual dots depends not only on physical factors such as the ejection method, nozzle size and mechanical tolerances but also on the algorithms implemented in the firmware of the printer.

\section{Data}
\label{section:data}
\begin{figure}[t]
\noindent
\begin{minipage}{0.3\linewidth}
\centering
\begin{tabular}{l|c}
        \toprule
         \textbf{Mfr.} & \textbf{\# models}  \\
         \midrule
         Canon & 8\\
         HP & 6\\
         Brother & 4\\
         Epson & 4\\
         Dell & 1\\
         Lexmark & 1\\
         Ricoh & 1\\
         \bottomrule
    \end{tabular}
\end{minipage}%
\hfill
\begin{minipage}{0.6\linewidth}
\centering
\includegraphics[width=.6\linewidth]{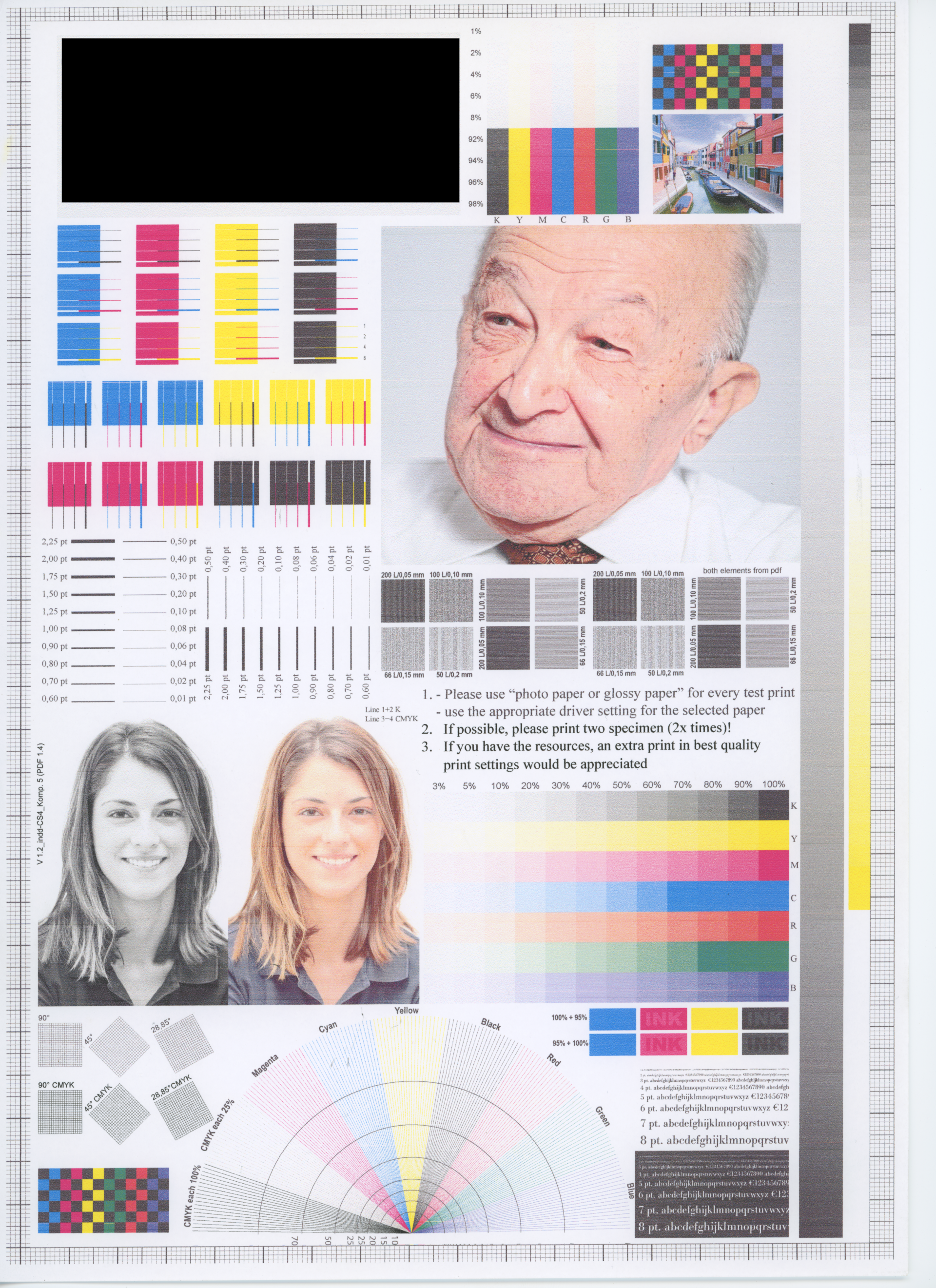}
\end{minipage}%
\caption{\textbf{Left:} Number of different printer models for each manufacturer present in the dataset. \textbf{Right:} Exemplar scan of the document template used in this dataset. The black area in the top left masks out printer model identifying text.}
\label{fig:sample}
\end{figure}

A dataset consisting of A4 scans of high resolution printed documents was collected with the help of various potential users of our system, leading to printing conditions that would occur in a real-world usage setting. In total, 25 unique printer models were used to print out a printing template (cf. Fig.~\ref{fig:sample}) containing varied content, such as text areas or portrait images. A list of all printer models contained in this dataset can be found in the experimental section in Fig.~\ref{fig:cm}, and per-manufacturer statistics in Fig.~\ref{fig:sample}. 

We gather two of such scans per printer model, arriving at 50 documents in total. One of the two scans is used for the training set, the other for validation purposes, allowing us to rule out possible model biases that focus on scan-induced variances such as micro-rotations. The printing quality and paper type were not controlled for and vary from document to document. 

A high resolution scan is necessary in order to observe inkjet droplets clearly, and as such we scan all documents with 2400 dpi using an "Epson V850 Pro" document scanner. We found that this resolution resulted in individually recognizable droplets and a generally good compromise between observed droplet quality and hardware requirements (cf. Fig.~\ref{fig:droplets}).

In order to minimize the influence that the global image content (such as it being a text document, a portrait image, etc.) has on the prediction, we do not directly extract features on the whole document, but on randomly sampled crops, each with a spatial size of $256\times256$ pixels, or $2.71 mm\times2.71 mm$. This is generally still large enough to observe complex droplet patterns, but separates it from the context of the document content. For each document, we sample 96 such crops and use these as basis for our feature extraction.

\subsection{Feature Extraction}
\begin{figure}[t]
    \centering
    \includegraphics[width=.5\textwidth]{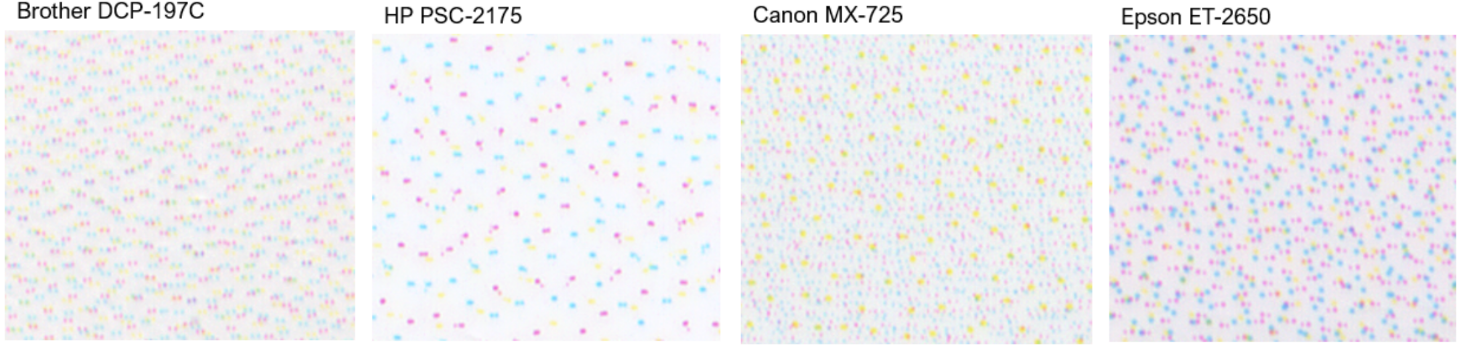}
    \caption{Close view of inkjet printed documents. It is immediately recognizable that individual printer models produce distinct droplet patterns.}
    \label{fig:droplets}
\end{figure}

Inkjet printers produce droplet patterns of a probabilistic nature. Upon taking a closer look (cf. Fig.~\ref{fig:droplets}) these patterns appear to be highly dependent on the printer model used and it leads to the question of what are the distinct factors that make up an individual printer model.

We decided to focus on two main aspects contained in these patterns: (1) The frequencies present in the image in order to capture the overall, global pattern of droplets, and (2) the droplet shape in order to capture detailed, local information. However, using these features directly would not be suitable as the feature vector would be too large and likely result in overfitting. As such, we compress them further by calculating their general statistics (cf. Tab.~\ref{tab:features}), which ideally removes sample-specific information, but keeps class-specific statistics intact. In the following we describe these features in more detail.

For the frequency domain features the wavelet transformation is of special interest, as it also considers the spatial aspect of the occurring frequencies. We found that Daubechies wavelets with three decomposition levels worked best. We then calculate the general statistics for each of the four resulting wavelet sub-bands individually.

For the droplet-specific features we first preprocess the image using general image processing techniques in order to make individual droplets more visible and distinct from the background. Image sharpening, denoising, and color thresholding helped in this regard. We then obtain individual droplets by extracting (closed) contours from the image crop and capture its shape by calculating the perimeter and area. We then aggregate these features over all droplets present in the crop by calculating their general statistics (cf. Tab.~\ref{tab:features}). These droplet features also allow easy filtering of unsuitable crops. If---for instance---a crop occurs in a plain area without any droplets it is immediately apparent from the features and an alternative crop location can be chosen.

Finally, we also add general image features that describe color and luminance by calculating (1) mean and standard deviation of pixel values, and (2) the contrast (based on the intensity channel Y of the YUV color representation). 

In our study of the printed documents we further observed that---depending on the printer model---each color channel also exhibited distinct droplet patterns. As such, we calculate the aforementioned features for each color channel individually (where applicable), in addition to features of a greyscale image variant which considers all droplet patterns at once.

In total, we arrive at 241 features and list them in Tab.~\ref{tab:features}. In order to facilitate smooth learning we further standardize all features individually according to their training statistics.

\begin{table}[t]
    \centering
    \caption{Overview of our extracted features in the case of wavelet-based frequency features.}
    \begin{tabular}{p{0.15cm} | p{3.5cm}}
        \toprule
        \multicolumn{2}{l}{\textbf{General Features}} \\
         \midrule
         \multicolumn{2}{l}{\textbf{Contrast}} \\
         &Max Intensity \\
         &Min Intensity \\
         &Contrast \\
         \midrule
         \multicolumn{2}{l}{\textbf{Per Color Channel Features} (C=4)} \\
         \midrule
         \multicolumn{2}{l}{\textbf{Colors}} \\
         &Pixel Mean \\
         &Pixel Std \\
         \multicolumn{2}{l}{\textbf{Droplet Area and Perimeter} (D=2)} \\
         &Mean \\
         &Std \\
         &25, 50, 75-th percentile \\
         \multicolumn{2}{l}{\textbf{Per Frequency Sub-Band Features} (L=4)} \\
         &Entropy \\
         &5, 25, 50, 75, 95-th percentile \\
         &Mean \\
         &Variance \\
         &Std \\
         &Root Mean Squared Value \\
         &\# Zero crossings \\
         &\# Mean crossings \\
         \bottomrule
    \end{tabular}
    \label{tab:features}
\end{table}

\section{Experiments}
\label{section:experiments}
\subsection{Setup}
For our classifier model we use a Multi-Layer Perceptron (MLP) with three hidden layers, each composed of 512 neurons and subsequent hyperbolic tangent activation functions. Among other classifier models we have found a neural network to be the most reliable (cf. Sec.~\ref{sec:ablations} for ablations).

The training process minimizes the cross-entropy loss using the Adam Optimizer~\cite{kingma_adam_2015} with an initial learning rate of $1e^{-4}$ and a L2 regularization loss with factor $1e^{-4}$. We train until the validation performance converges.

We measure the classification performance based on the F1-Score. We also report its Top-N performance for $N=2\ldots5$, in which we consider a prediction to be a true positive if it is within the top $N$ predictions of the classifier. All scores are averaged over three different random seeds and we also report the resulting standard deviation.

We compare different frequency-based features and also state-of-the-art image classification models which are tailored for limited amounts of data, such as MobileNet-V3~\cite{howard_searching_2019}, ResNet-18~\cite{he_deep_2015}, and SqueezeNet~\cite{iandola_squeezenet_2016}.

\subsection{Printer Classification Baseline}
Here we analyze the baseline printer classification performance. We compare different kinds of frequency-based features and also compare with image-based classifiers. For each method for domain transfer into frequency space, we perform the extraction on the two-dimensional image (termed 2D in the tables), as well as a row-wise linearized version that calculates the features for each row of pixels individually and concatenates them afterwards.
We show the results in Tab.~\ref{tab:base_performance}.

\begin{table}[t]
    \caption{Baseline performance for different frequency-based features and also image-based models on our dataset.}
    \label{tab:base_performance}
    \vskip 0.1in
    \begin{center}
    \resizebox{.48\textwidth}{!}{%
    \begin{tabular}{p{1.7cm}|l|l|l|l|l}
\toprule
                      & \multicolumn{5}{c}{\textbf{F1}~$\uparrow$}\\
                      & \textbf{Top1} & \textbf{Top2} & \textbf{Top3} & \textbf{Top4} & \textbf{Top5} \\\midrule
                    \textbf{Wavelets}         &  $32.8$\textcolor{darkgray}{\scriptsize$\pm0.6$} &  $62.2$\textcolor{darkgray}{\scriptsize$\pm0.3$} &  $69.4$\textcolor{darkgray}{\scriptsize$\pm0.9$} &  $75.6$\textcolor{darkgray}{\scriptsize$\pm1.1$} &  $80.0$\textcolor{darkgray}{\scriptsize$\pm1.1$} \\
                    Wavelets-2D &  $31.2$\textcolor{darkgray}{\scriptsize$\pm0.6$} &  $61.0$\textcolor{darkgray}{\scriptsize$\pm0.5$} &  $69.9$\textcolor{darkgray}{\scriptsize$\pm0.6$} &  $75.4$\textcolor{darkgray}{\scriptsize$\pm0.4$} &  $79.6$\textcolor{darkgray}{\scriptsize$\pm0.6$} \\
                    STFT-2D     &  $29.1$\textcolor{darkgray}{\scriptsize$\pm1.0$} &  $60.0$\textcolor{darkgray}{\scriptsize$\pm0.7$} &  $69.0$\textcolor{darkgray}{\scriptsize$\pm0.9$} &  $74.9$\textcolor{darkgray}{\scriptsize$\pm0.4$} &  $80.2$\textcolor{darkgray}{\scriptsize$\pm0.2$} \\
                    FFT-2D &  $29.0$\textcolor{darkgray}{\scriptsize$\pm0.4$} &  $60.9$\textcolor{darkgray}{\scriptsize$\pm0.5$} &  $69.5$\textcolor{darkgray}{\scriptsize$\pm0.2$} &  $75.4$\textcolor{darkgray}{\scriptsize$\pm0.2$} &  $79.7$\textcolor{darkgray}{\scriptsize$\pm0.6$} \\
                    STFT   &  $28.3$\textcolor{darkgray}{\scriptsize$\pm0.4$} &  $57.5$\textcolor{darkgray}{\scriptsize$\pm0.4$} &  $66.8$\textcolor{darkgray}{\scriptsize$\pm0.4$} &  $73.3$\textcolor{darkgray}{\scriptsize$\pm0.1$} &  $78.3$\textcolor{darkgray}{\scriptsize$\pm0.2$} \\
                    FFT    &  $25.4$\textcolor{darkgray}{\scriptsize$\pm0.9$} &  $54.5$\textcolor{darkgray}{\scriptsize$\pm0.3$} &  $63.2$\textcolor{darkgray}{\scriptsize$\pm0.4$} &  $69.3$\textcolor{darkgray}{\scriptsize$\pm0.5$} &  $74.3$\textcolor{darkgray}{\scriptsize$\pm0.2$} \\

                    \midrule
ResNet-18    &  $18.3$\textcolor{darkgray}{\scriptsize$\pm2.0$} &  $44.7$\textcolor{darkgray}{\scriptsize$\pm2.7$} &  $54.8$\textcolor{darkgray}{\scriptsize$\pm2.2$} &  $62.5$\textcolor{darkgray}{\scriptsize$\pm2.0$} &  $68.4$\textcolor{darkgray}{\scriptsize$\pm1.4$} \\
SqueezeNet   &  $17.6$\textcolor{darkgray}{\scriptsize$\pm1.7$} &  $45.4$\textcolor{darkgray}{\scriptsize$\pm1.9$} &  $56.9$\textcolor{darkgray}{\scriptsize$\pm2.8$} &  $65.2$\textcolor{darkgray}{\scriptsize$\pm2.1$} &  $72.2$\textcolor{darkgray}{\scriptsize$\pm1.4$} \\
MobileNet-V3 &   $4.0$\textcolor{darkgray}{\scriptsize$\pm0.0$} &  $14.8$\textcolor{darkgray}{\scriptsize$\pm0.0$} &  $21.5$\textcolor{darkgray}{\scriptsize$\pm0.2$} &  $27.6$\textcolor{darkgray}{\scriptsize$\pm0.0$} &  $33.0$\textcolor{darkgray}{\scriptsize$\pm0.4$} \\
\bottomrule
\end{tabular}
    }
    \end{center}
    \vskip -0.1in
\end{table}

All frequency-based features are able to significantly outperform image-based models. Among those, the 1D wavelet features show the best Top-1 F1-Score. While both 1D and 2D-based frequency features do not show large differences for each method, the locality information present in the STFT approach and especially in the Wavelet features seems to be beneficial for overall performance. 

For all approaches the Top-2 F1-Score is roughly twice as high as the Top-1 performance, indicating that the model ranks the true printer model high, but often not high enough. In the following we provide a possible solution to mitigate this issue.

\subsection{Averaging Crop Predictions}
While the prediction performance of our model is substantially better than image-based predictors, the majority of samples is still classified wrong. One document scan of a given printer is represented in our dataset as multiple crop samples, which are handled independently of each other. As we can assume that the whole document is available at inference time, we also investigate the per-document prediction performance by aggregating the predictions of all crops of a given document. We achieve this by simply averaging the resulting network logits before calculating the metrics. As can be seen in Tab.~\ref{tab:aggregated_performance}, the resulting per-document classification performance is significantly higher than the per-crop performance, without any changes to the training process of the model. 

\begin{table}[t]
    \caption{Per-document prediction performance of our approach and relevant related work. The scores increase across the board for almost all models compared to the per-crop performance seen in Tab.~\ref{tab:base_performance}.}
    \label{tab:aggregated_performance}
    \vskip 0.1in
    \begin{center}
    \resizebox{.48\textwidth}{!}{%
    \begin{tabular}{p{1.7cm}|l|l|l|l|l}
\toprule
                      & \multicolumn{5}{c}{\textbf{F1}~$\uparrow$}\\
                      & \textbf{Top1} & \textbf{Top2} & \textbf{Top3} & \textbf{Top4} & \textbf{Top5} \\
                    \midrule
                    \textbf{Wavelets}         &  $57.3$\textcolor{darkgray}{\scriptsize$\pm2.3$} &  $77.0$\textcolor{darkgray}{\scriptsize$\pm1.8$} &   $79.0$\textcolor{darkgray}{\scriptsize$\pm1.7$} &  $87.2$\textcolor{darkgray}{\scriptsize$\pm1.5$} &  $88.0$\textcolor{darkgray}{\scriptsize$\pm1.5$} \\
STFT        &  $57.3$\textcolor{darkgray}{\scriptsize$\pm2.3$} &  $78.0$\textcolor{darkgray}{\scriptsize$\pm0.0$} &  $78.0$\textcolor{darkgray}{\scriptsize$\pm0.0$} &  $80.0$\textcolor{darkgray}{\scriptsize$\pm1.7$} &  $86.3$\textcolor{darkgray}{\scriptsize$\pm2.6$} \\                    
FFT         &  $57.3$\textcolor{darkgray}{\scriptsize$\pm2.3$} &  $72.9$\textcolor{darkgray}{\scriptsize$\pm1.9$} &  $77.0$\textcolor{darkgray}{\scriptsize$\pm1.8$} &  $87.2$\textcolor{darkgray}{\scriptsize$\pm1.5$} &  $88.9$\textcolor{darkgray}{\scriptsize$\pm2.5$} \\
STFT-2D     &  $52.0$\textcolor{darkgray}{\scriptsize$\pm0.0$} &  $73.9$\textcolor{darkgray}{\scriptsize$\pm3.6$} &  $80.0$\textcolor{darkgray}{\scriptsize$\pm1.7$} &  $86.3$\textcolor{darkgray}{\scriptsize$\pm2.6$} &  $90.5$\textcolor{darkgray}{\scriptsize$\pm1.4$} \\
FFT-2D      &  $54.7$\textcolor{darkgray}{\scriptsize$\pm2.3$} &  $76.0$\textcolor{darkgray}{\scriptsize$\pm1.8$} &  $79.9$\textcolor{darkgray}{\scriptsize$\pm3.3$} &  $87.2$\textcolor{darkgray}{\scriptsize$\pm1.5$} &  $88.9$\textcolor{darkgray}{\scriptsize$\pm0.0$} \\
Wavelets-2D &  $50.7$\textcolor{darkgray}{\scriptsize$\pm2.3$} &  $77.0$\textcolor{darkgray}{\scriptsize$\pm1.8$} &  $81.9$\textcolor{darkgray}{\scriptsize$\pm1.6$} &  $83.7$\textcolor{darkgray}{\scriptsize$\pm0.0$} &  $86.3$\textcolor{darkgray}{\scriptsize$\pm2.6$} \\
                    \midrule
ResNet-18    &  $36.0$\textcolor{darkgray}{\scriptsize$\pm4.0$} &  $68.0$\textcolor{darkgray}{\scriptsize$\pm9.5$} &  $76.5$\textcolor{darkgray}{\scriptsize$\pm10.2$} &  $79.5$\textcolor{darkgray}{\scriptsize$\pm9.7$} &  $81.5$\textcolor{darkgray}{\scriptsize$\pm8.4$} \\
SqueezeNet   &  $33.3$\textcolor{darkgray}{\scriptsize$\pm2.3$} &  $66.0$\textcolor{darkgray}{\scriptsize$\pm4.2$} &   $69.5$\textcolor{darkgray}{\scriptsize$\pm1.9$} &  $78.0$\textcolor{darkgray}{\scriptsize$\pm3.0$} &  $83.7$\textcolor{darkgray}{\scriptsize$\pm2.7$} \\
MobileNet-V3 &   $4.0$\textcolor{darkgray}{\scriptsize$\pm0.0$} &  $14.8$\textcolor{darkgray}{\scriptsize$\pm0.0$} &   $21.4$\textcolor{darkgray}{\scriptsize$\pm0.0$} &  $27.6$\textcolor{darkgray}{\scriptsize$\pm0.0$} &  $33.3$\textcolor{darkgray}{\scriptsize$\pm0.0$} \\
\bottomrule
\end{tabular}
    }
    \end{center}
    \vskip -0.1in
\end{table}

Aggregating the crops into a single document can, however, happen at different points in the model pipeline. In Tab.~\ref{tab:base_performance} we have shown the case where individual crops are fed to the model independently of each other, and the aggregation happened afterwards. This aggregation could also happen at the feature side, where feature vectors from crops belonging to a single document are averaged before being put through the model, resulting in a single per-document prediction. We investigate this configuration and also distinguish whether this aggregation happens during training and / or during validation of the model in Tab.~\ref{tab:dataset_setup_comp}. These results indicate that the best approach is to train on individual crops, and validate on averaged model logits.

\begin{table}[t]
    \caption{Prediction performance of our model for different training and validation configurations. The crop aggregation column refers to the point at which individual crops of a single document are combined into a single document entity for classification purposes.}
    \label{tab:dataset_setup_comp}
    \vskip 0.1in
    \begin{center}
    \resizebox{.48\textwidth}{!}{%
    \begin{tabular}{p{2.5cm}|p{1.5cm}|p{1.5cm}|l}
\toprule
                      \textbf{Crop Aggregation} & \textbf{Training Aggregation} & \textbf{Validation Aggregation} & \textbf{F1-Top1}~$\uparrow$\\
                      \midrule
                           After Prediction & No & Yes  &  $57.3$\textcolor{darkgray}{\scriptsize$\pm2.3$} \\
                           After Prediction & Yes & Yes & $37.3$\textcolor{darkgray}{\scriptsize$\pm10.1$} \\
                    None         & No & No &  $32.8$\textcolor{darkgray}{\scriptsize$\pm0.6$} \\
                    Before Prediction & Yes & Yes & $32.0$\textcolor{darkgray}{\scriptsize$\pm0.0$} \\
                    Before Prediction & Yes & No & $8.3$\textcolor{darkgray}{\scriptsize$\pm0.2$} \\
\bottomrule
\end{tabular}
    }
    \end{center}
    \vskip -0.1in
\end{table}

\begin{figure}[t]
    \centering
    \includegraphics[width=.5\textwidth]{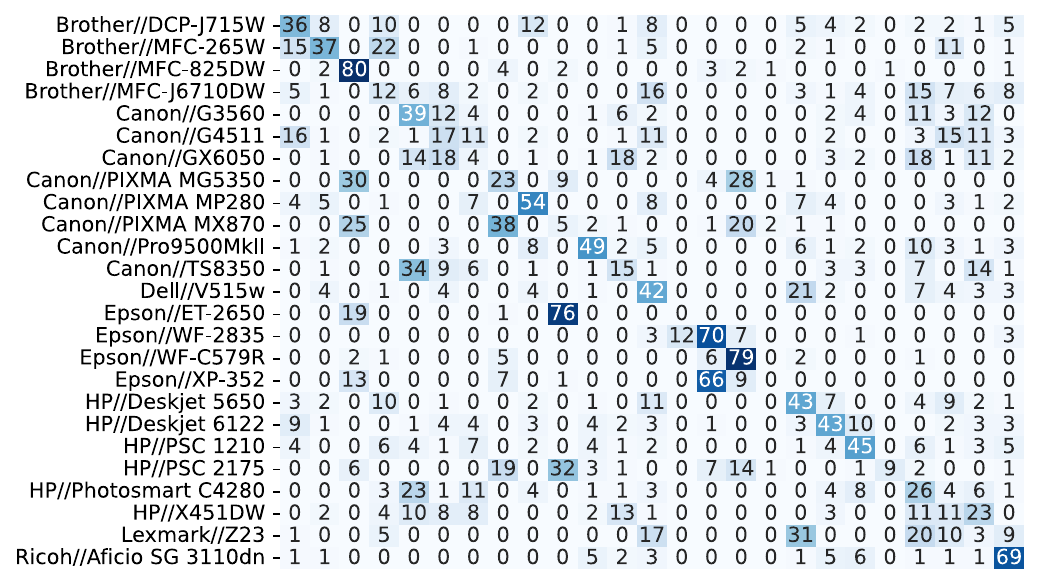}
    \includegraphics[width=.5\textwidth]{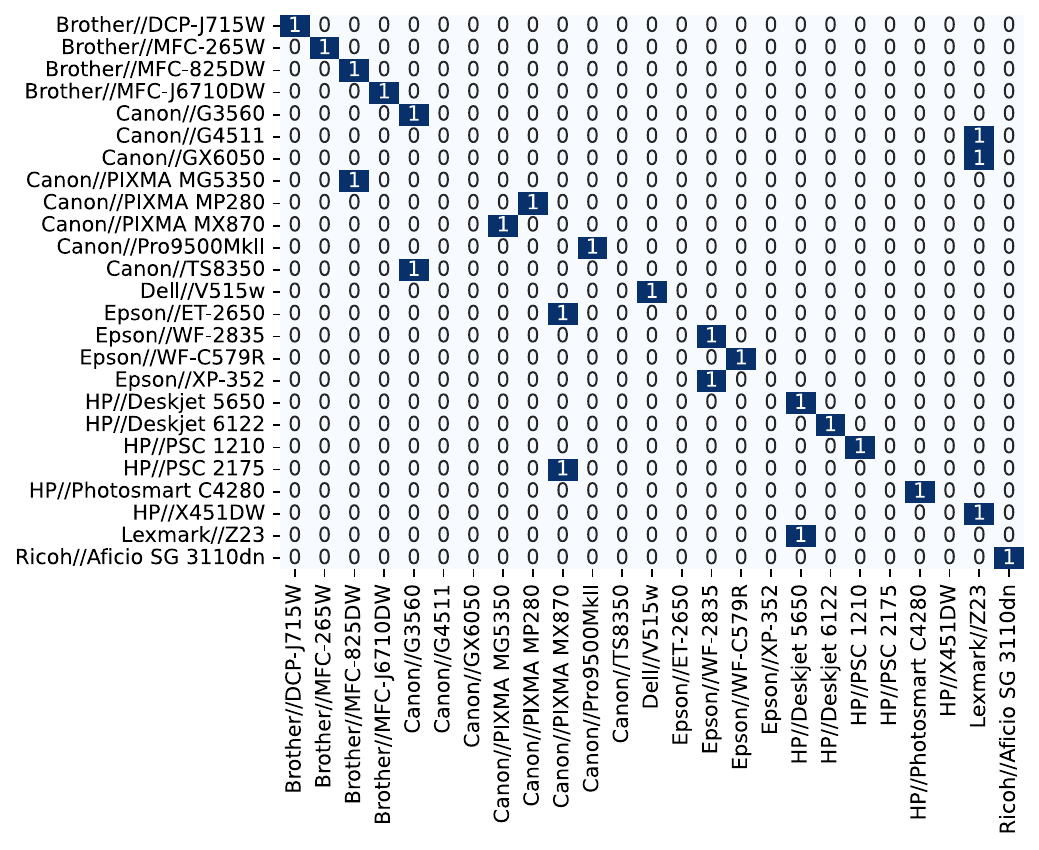}
    \caption{Confusion Matrices for our prediction model for per-crop predictions (Top) and per-document predictions (Bottom). The rows describe the true classes, while the columns refer to the predicted classes.}
    \label{fig:cm}
\end{figure}

Overall, the improved per-document performance is intriguing, as it indicates that---on average---individual crop predictions are more likely to be wrong than their aggregation. The confusion matrix shown in Fig.~\ref{fig:cm} gives an overview of the miss-classifications. It is immediately recognizable that many classes (such as for instance ``Brother//MFC-J6710DW'') for which the sum of wrong crop predictions surpasses the amount of true positives in the diagonal are still assigned correctly in the per-document classification. Since we average the network logits and not the predictions in the per-document classification setting, this indicates that in many cases the model assigns a high logit to the true class, even if it is not the highest, and their average then surpasses all other class logits. This also explains the significant performance gain when considering the Top-2 F1 performance compared to the Top-1, where the true class is often only the second highest logit.

\subsection{Further Ablations}\label{sec:ablations}
Here we investigate further ablations related to the features and the classifier. For model ablations we use the 1D wavelet features as basis, and choose the following machine learning approaches: (1) Support-Vector-Machine (SVM), (2) Random Forest (N=500 trees), and (3) XGBoost (N=500 trees). For the feature variations we use our default MLP classifier model and analyze the following setups: (1) We do not extract per color channel features and instead only use a grey-scale image as basis, and (2) we do not use features besides the frequency statistics. We show the results in Tab.~\ref{tab:ablations}.

\begin{table}[t]
    \caption{Further model- and feature-related ablations of our proposed approach.}
    \label{tab:ablations}
    \vskip 0.1in
    \begin{center}
    \begin{tabular}{p{0.2cm}p{3.5cm}|l|l}
\toprule
                      && \multicolumn{2}{c}{\textbf{F1-Top1}~$\uparrow$}\\
                      && Per-Crop & Per-Document \\
                    \midrule
                    \multicolumn{2}{l|}{\textbf{Reference}} & $32.8$\textcolor{darkgray}{\scriptsize$\pm0.6$} & $57.3$\textcolor{darkgray}{\scriptsize$\pm2.3$} \\
                    \midrule
                    \multicolumn{2}{l|}{\textbf{Model Variations}} & & \\
                    &Random Forest         & $32.2$\textcolor{darkgray}{\scriptsize$\pm0.5$} & $52.0$\textcolor{darkgray}{\scriptsize$\pm0.0$}\\
                    &SVM       & $23.3$\textcolor{darkgray}{\scriptsize$\pm0.0$}  & $54.7$\textcolor{darkgray}{\scriptsize$\pm2.3$} \\
                    &XGBoost & $19.6$\textcolor{darkgray}{\scriptsize$\pm0.0$} & $56.0$\textcolor{darkgray}{\scriptsize$\pm0.0$} \\
                    \midrule
                    \multicolumn{2}{l|}{\textbf{Feature Variations}} & &\\
                    &Single Color Channel Features        & $27.0$\textcolor{darkgray}{\scriptsize$\pm0.2$} & $58.7$\textcolor{darkgray}{\scriptsize$\pm2.3$}\\
                    &Only Frequency Features         & $29.3$\textcolor{darkgray}{\scriptsize$\pm0.5$} & $58.7$\textcolor{darkgray}{\scriptsize$\pm2.3$}\\
\bottomrule
\end{tabular}
    \end{center}
    \vskip -0.1in
\end{table}

Among the tested model variations, the random forest approach was able to achieve a similar per-crop performance as our MLP classifier, albeit at a lower per-document performance. On the other hand, the XGBoost method achieves comparable per-document performance, but much lower per-crop performance. For the feature variations the performance on the per-document evaluation is stable, however for the per-crop performance a combination of all features as proposed led to the best result.

\section{Limitations and Future Work}
\label{section:limitations}
While we have shown that manually extracted frequency-based features are better than relying on image features directly, there is still room for further improvements. Especially in the area of forensics a high level of confidence and reliance is necessary in order to integrate such a system in the workflow. One aspect that makes the whole endeavour challenging in general is the difficulty in obtaining large amounts of data that correspond to a wide variety of printing conditions, however our dataset can be a suitable basis in this regard to expand upon. For this reason and due to the large amount of different printer models in the real world, it is highly likely that a tested document is from a printer model not available in the dataset. As such, model extensions that can predict such outliers is a promising direction for future work that makes this more applicable.

Furthermore, in this work we have analyzed image-based features and our manually extracted features separately. It might be even better to fuse both types of features in a single model to also let the model extract useful features on its own. These could then potentially replace the droplet-based features that we utilized. However at this time the limiting factor here seems to be the small amount of data available. As such, a clear future direction is the expansion of the dataset both in terms of document content variety and also new printer models. We reckon that, as in other areas, deep learning methods will gradually become more important the larger the data basis gets.

An interesting other direction for future work might be the identification of not only the printer model used, but also the printer instance. The hypothesis is that due to mechanical wear or slight manufacturing differences the droplet dispersion pattern varies even among same printer models. It would be interesting to pursue whether this can be detected from the features that we proposed as well. Similarly, the used scanner device might also be an important factor that should be taken into account. In our dataset we used only a single scanning device, in reality however this is difficult to enforce. Therefore analyzing the performance on documents scanned with different scanning devices might provide valuable insights.

\section{Conclusion}
\label{section:conclusion}
We introduced a novel dataset and corresponding training and evaluation scheme for printer model identification that works without specialized scanning hardware. We proposed a set of manually extracted features which lead to substantial performance improvements over relying on image features alone, and show that wavelet-based frequency features best capture the printer identities. In the future this work will be part of a system to be used by document forensics experts.

\section*{Acknowledgment}
We extend our gratitude to the Document Forensics Department of the Baden Württemberg State Office of Criminal Investigations for their invaluable assistance in providing the printed documents essential to this project. Special recognition is owed to the department's head, Rolf Fauser, whose personal dedication and insightful expertise were instrumental in facilitating our work.

\bibliographystyle{./IEEEtran}
\bibliography{./IEEEabrv,./ict.bib}

\end{document}